\documentclass[conference]{IEEEtran}
\IEEEoverridecommandlockouts
\usepackage[pscoord]{eso-pic}
\usepackage[mathlines,switch]{lineno}

\makeatletter

\let\old@ps@IEEEtitlepagestyle\ps@IEEEtitlepagestyle
\def\confheader#1{%
    \def\ps@IEEEtitlepagestyle{%
        \old@ps@IEEEtitlepagestyle%
        \def\@oddhead{\strut\hfill#1\hfill\strut}%
        \def\@evenhead{\strut\hfill#1\hfill\strut}%
    }%
    \ps@headings%
}
\makeatother

\usepackage{cite}
\usepackage{amsmath,amssymb,amsfonts}
\usepackage{algorithmic}
\usepackage{graphicx}
\usepackage{textcomp}
\usepackage{xcolor}
\usepackage{url}
\usepackage[export]{adjustbox}
\def\BibTeX{{\rm B\kern-.05em{\sc i\kern-.025em b}\kern-.08em
    T\kern-.1667em\lower.7ex\hbox{E}\kern-.125emX}}
\begin{document}

\title{Subjective Question Generation and Answer Evaluation using NLP\\
}

\author{\IEEEauthorblockN{G. M. Refatul Islam}
\IEEEauthorblockA{\textit{School of Data and Sciences} \\
\textit{Brac University}\\
Dhaka, Bangladesh \\
gm.refatul.islam@g.bracu.ac.bd}
\and
\IEEEauthorblockN{Safwan Shaheer}
\IEEEauthorblockA{\textit{School of Data and Sciences} \\
\textit{Brac University}\\
Dhaka, Bangladesh \\
safwan.shaheer@bracu.ac.bd}
\and
\IEEEauthorblockN{Yaseen Nur}
\IEEEauthorblockA{\textit{School of Data and Sciences} \\
\textit{Brac University}\\
Dhaka, Bangladesh \\
yaseen.nur.taz@bracu.ac.bd}
\and
\IEEEauthorblockN{Mohammad Rafid Hamid}
\IEEEauthorblockA{\textit{School of Data and Sciences} \\
\textit{Brac University}\\
Dhaka, Bangladesh \\
mohammad.rafid.hamid@g.bracu.ac.bd}

}

\maketitle

\begin{abstract}
Natural Language Processing (NLP) is one of the most revolutionary technologies today. It uses artificial intelligence to understand human text and spoken words. It is used for text summarization, grammar checking, sentiment analysis, and advanced chatbots and has many more potential use cases. Furthermore, it has also made its mark on the education sector. Much research and advancements have already been conducted on objective question generation; however, automated subjective question generation and answer evaluation are still in progress. An automated system to generate subjective questions and evaluate the answers can help teachers assess student work and enhance the student's learning experience by allowing them to self-assess their understanding after reading an article or a chapter of a book. This research aims to improve current NLP models or make a novel one for automated subjective question generation and answer evaluation from text input.
\end{abstract}


\section{Introduction}

In the dynamic landscape of educational technology, our research primarily addresses a significant and often overlooked challenge: the subjective evaluation of student responses, especially in the context of higher-order thinking skills as categorized by Bloom's Taxonomy. While previous studies have predominantly focused on objective question generation and evaluation, our work takes a groundbreaking step by not only generating subjective questions but also by meticulously evaluating the answers to these questions. \textbf {This paper presents an innovative approach that harnesses the advanced capabilities of Large Language Models (LLMs) for both generating nuanced subjective questions and conducting sophisticated evaluations of student responses.}

Our research stands out by emphasizing the subjective answer evaluation aspect, arguably a more complex and impressive task than question generation. We detail a comprehensive framework that outlines the methods for generating data suitable for both question generation and answer evaluation at the higher levels of Bloom's Taxonomy. The crux of our study lies in the instruct-tuning of large language models, which are adept at understanding and processing the complexities inherent in subjective answers. This approach not only bridges the gap in existing educational technologies but also sets a new precedent in the automated assessment of higher-order cognitive skills, a domain traditionally dominated by human evaluators. The paper thoroughly analyzes the efficacy of these models, highlighting their potential to revolutionize the way subjective understanding is measured and fostered in educational settings.

\section{Related Works}
\label{background}

\subsection{Question Generation}
Deena et al. \cite{2}, Du et al. \cite{3}, and Klein et al. \cite{4} explore automatic question generation. Deena et al. focus on e-learning, using Bloom's Taxonomy and Named Entity Recognizer. Du et al. adopt a neural network-based methodology, employing an RNN Encoder-Decoder architecture for the purpose of answering reading comprehension questions. Klein et al. combines the capabilities of GPT-2 and BERT to create a system called "Learning to Answer by Learning to Ask" that integrates question creation and answering.
Additional research conducted by Mohd et al. \cite{13}, and Agarwal et al. \cite{14} focuses on the characteristics of optimal survey questions and the utilization of discourse connectives to generate more impactful questions.  

\subsection{Answer Evaluation}
Dhokrat et al. \cite{1}, Sakaguchi et al. \cite{7}, and Loukina et al. \cite{8} present varied approaches for evaluating answers. Dhokrat et al. present a system that employs Natural Language Processing (NLP) approaches to handle subjective queries. Sakaguchi et al. integrate response-based and reference-based approaches to evaluate brief answers, whereas Loukina et al. employ a combination of natural language and speech-processing methods to assess English proficiency.
In addition, Hou et al. \cite{10}, Madnani et al. \cite{11}, and Alrehily et al. \cite{15} have made contributions to the field by creating systems that automatically evaluate student answers, assess reading comprehension questions, and compare student responses to the instructor's reference replies, respectively.  

\subsection{LLMs}
Research papers such as LLAMA \cite{touvron2023llama}, InstructGPT \cite{ouyang2022training}, and Self-Instruct \cite{wang2023selfinstruct} primarily concentrate on enhancing the training and fine-tuning of language models. The effectiveness of smaller models trained on huge datasets is demonstrated by LLAMA. InstructGPT utilizes instruct-tuning and reinforcement learning to enhance its ability to follow instructions, whereas Self-Instruct diminishes the reliance on human involvement in training by producing novel tasks.

Further progress is observed in Alpaca \cite{alpaca} and LIMA \cite{lima}, which optimize models for particular tasks, and in LoRA \cite{hu2021lora} and QLORA \cite{qlora}, which introduce effective approaches for optimizing models. Mistral 7B \cite{jiang2023mistral} further showcases the effectiveness of well-crafted, smaller models.

\section{Research Methodology}
\label{method}

\begin{figure}
    \centering
    \includegraphics[width=.24\textwidth]{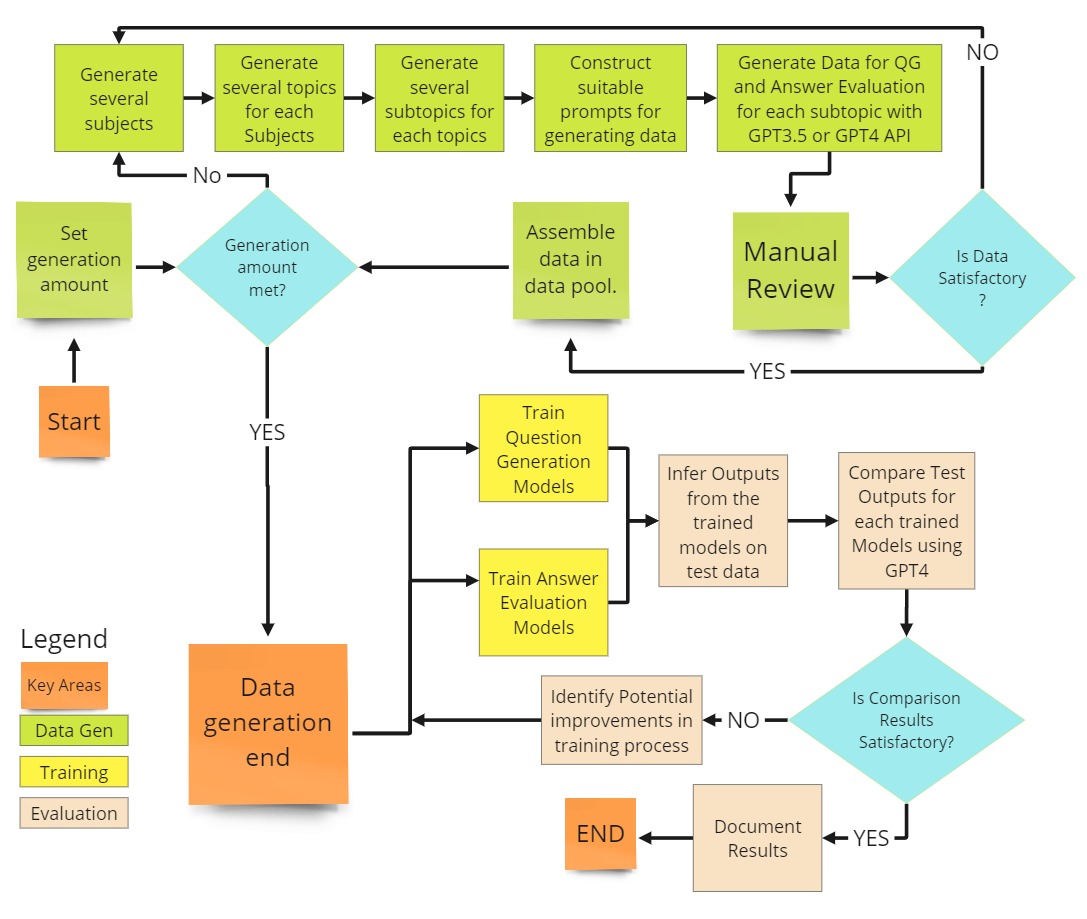}
    \caption{Research Workflow}
    \label{fig:research-workflow}
\end{figure}

\subsection{Data}
As we mentioned earlier, there has been very little research on subjective questions. Hence, no proper dataset existed for subjective question generation and answer evaluation. As a result, we generated synthetic data for our study. We generated two datasets consisting of 6978 and 4523 samples for subjective question generation and answer evaluation, respectively. 

For question generation, each sample data consisted of \textbf{id}, \textbf{subject}, \textbf{topic}, \textbf{subtopic}, \textbf{context}, \textbf{question}. For the evaluation dataset, each sample contained \textbf{id}, \textbf{question}, \textbf{evaluation criteria}, \textbf{student answer}, \textbf{answer quality}, \textbf{answer evaluation}, \textbf{grammar score}, \textbf{coherence score}, \textbf{relevance score}.

\subsection{Data Generation}
The procedure for generating both datasets was meticulously orchestrated, adopting a uniform and methodical approach inspired by the strategies employed in creating their dataset for the Alpaca \cite{alpaca}.

For a generation, we made use of the GPT-4 API. We manually curated a list of subjects, topics, and subtopics for subjective question generation. Each subject contained a set of topics and had a set of subtopics. This step was to ensure that there would be no duplicate question generation. In the prompt, we randomly assigned a subject, topic, and subtopic and asked GPT-4 to generate a context passage and a question from each of the upper three Bloom's Taxonomy Levels (Analysis, Synthesis, Evaluation) from the generated context.

On the other hand, For answer evaluation, we have prepared a distinctive dataset consisting of 4,523
unique datums, specifically developed for answer evaluation. The data structure for
this is also similar to question generation model. As there were several input output
fields we decided to keep things seperate. 
The dataset included fields like \textbf{SubjectiveQuestion}, \textbf{EvaluationCriteria}, \textbf{studentAnswer} as inputs, and \textbf{answerEvaluation}, \textbf{score}, and \textbf{type} as outputs.The scoring system evaluated grammar, coherence, and relevance. The 'type' field categorized the correctness of the student's answer into five levels: \textbf {Perfect, Moderate, Average, BelowAverage, and Imperfect.}

\subsection{Models}
LLaMA 7B is an auto-regressive language model with 7 billion total parameters. It was trained on 1 trillion tokens and has a vocabulary size of 32,000. The model was trained from January 2023 to July 2023 and is licensed under a non-commercial bespoke license \cite{touvron2023llama}

Mistral 7B, on the other hand, has 7.3 billion total parameters. It uses Grouped-query Attention and Sliding Window Attention in its architecture and is licensed under Apache-2.0 \cite{almazrouei2023falcon}

Falcon 7B is a causal decoder-only model with 7 billion total parameters. It was trained on 1,500 billion tokens and has a total of 71 heads. This model is also licensed under Apache-2.0 \cite{jiang2023mistral}

\subsection{Training}

In our research, we trained three large language models: Llama 7b, Falcon 7b, and Mistral 7b, using a single NVIDIA RTX A6000 GPU. We made use of lit-gpt \cite{lit-gpt-2023} for the training and evaluation process. Our training process involved an 80/15/5 stratified split for training, validation, and testing based on the question and answer type. Even though all the models had 7b parameters, due to QLoRA, only 21 million parameters were fine-tuned. The training took roughly ~5 hrs for the answer evaluation model and ~3.5 hrs for the question generation model.
\\

\subsubsection{Instruct Tuning}
Instruct tuning \cite{ouyang2022training} was employed as a valuable approach for refining transformer models to better adhere to directions. This technique was particularly useful for tasks that involve producing responses based on explicit instructions, such as subjective question creation and answer evaluation. We leveraged our self-instruct dataset, similar to the Stanford team's approach in developing their Alpaca model \cite{alpaca}, to fine-tune LLaMA into an instruction-following model effectively.
\\

\subsubsection{LoRA}
LoRA, a Low-Rank Adaptation technique, was used as a cost-effective and efficient way to fine-tune our Large Language Models. This method reduced the number of trainable parameters, improved memory efficiency, simplified task switching, and provided comparable performance to full fine-tuning \cite{hu2021lora}
\\

\subsubsection{4-bit NormalFloat Quantization}
4-bit NormalFloat (NF4) quantization is a novel encoding optimized for the distribution of neural network weights. It compresses the full pre-trained language model to reduce memory requirements. The quantized base model is then supplemented with low-rank adapters added densely throughout the layers. The adapters use full 16-bit precision and are fine-tuned while the base model remains fixed. \\

\subsubsection{Bfloat16 Floating Point Precision}
Bfloat16 is a 16-bit floating point format that reduces precision from 24 bits to 8 bits while preserving the number range of the 32-bit IEEE 754 single-precision floating-point format (binary32). This format is frequently utilized in mixed-precision arithmetic since it allows bfloat16 numbers to be expanded to larger data types and acted upon. It is particularly useful in machine learning and deep learning applications due to its balance between range and precision. \\

\subsubsection{QLoRA}
Lastly, we employed QLoRA \cite{qlora}, a technique that uses a high-precision method to quantize a pre-trained model to 4 bits, followed by the integration of a compact set of learnable Low-Rank Adapter weights that are fine-tuned with the backpropagation step through the quantized weights. This approach balanced precision and computation, making it a viable method for fine-tuning large language models.\\

\subsubsection{Hyperparameters}
The training was configured with the following hyperparameters:\\

\begin{itemize}
  \item Learning rate: $2 \times 10^{-4}$
  \item Batch size: 128
  \item Micro batch size: 1
  \item Gradient accumulation iterations: batch\_size
  \item Maximum sequence length: 2500
  \item Maximum iterations: 10,000 (train dataset size)
  \item Weight decay: 0.01
  \item Quantization: bnb.nf4
  \item Precision: bfloat16 floating-point format
  \item LoRA parameters:
  \begin{itemize}
    \item $r$: 8
    \item $\alpha$: 16
    \item Dropout: 0.1
    \item Query, Key, Value, Projection, MLP, Head: Enabled
  \end{itemize}
  \item Warmup steps: 100\\
\end{itemize}

\subsubsection{Training Input Prompt \& Output}

During training, we provided a custom prompt as input to our models based on the task. The prompt consisted of a general instruction asking it to follow the provided task's instructions, as performed by the Stanford alpaca team \cite{alpaca}. After that, we provided the task-specific sections of our prompt. We provided the subjective question, the evaluation criteria to use as a rubric, and the student's answers for subject question evaluation. As for the output, the model provided the evaluation and the three scores for relevance, coherence, and grammar.

Similarly, we used a specific prompt for the question generation part. In this prompt, we only provided the generic instruction to generate a specific Bloom's taxonomy question and the passage from which the question has to be generated. The model's output was the question.\\

\subsubsection{Training Process}
\begin{enumerate}
\item The code begins by setting up hyperparameters such as learning rate, batch size, and weight decay, among others. These parameters are crucial for controlling the learning process of the model.
\item We then set up the training environment, including the data directory, checkpoint directory, and output directory. It also sets up the precision and quantization for the training process.
\item The model is loaded from a checkpoint, and only the LoRA layers are marked as trainable. This is done to fine-tune the model on a specific task.
\item The optimizer used is AdamW, which is a variant of the Adam optimizer that includes weight decay. This optimizer is known for its efficiency and effectiveness in training deep learning models.
\item A learning rate scheduler is used, specifically the CosineAnnealingLR scheduler. This scheduler adjusts the learning rate according to a cosine function, which can help achieve a better model by exploring different learning rates during training.
\item The training process involves iterating over the training data 
 by providing the training prompt to the model, generating output, computing the loss, and updating the model parameters using the optimizer. The loss function used is cross-entropy.
\item The model is evaluated at regular intervals during training. This involves computing the loss on a validation set and generating some output from the model to check its performance.
\item Finally, the trained model is saved for future use. The saving process involves filtering out the LoRA weights, which are the only parts of the model that have been updated during training.\\
\end{enumerate}

\begin{figure}
    \centering
    \includegraphics[width=.24\textwidth]{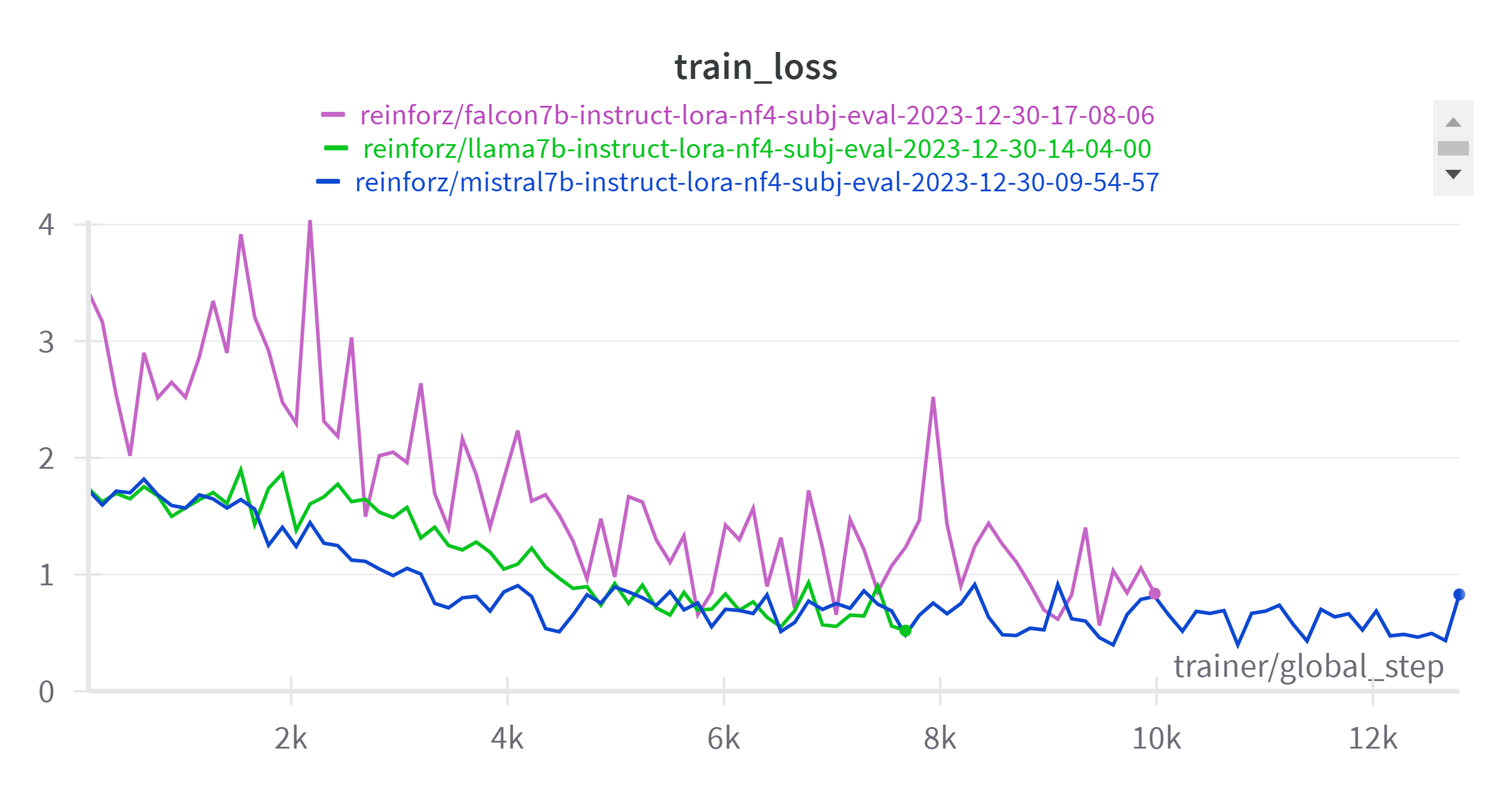}\hfill
    \includegraphics[width=.24\textwidth]{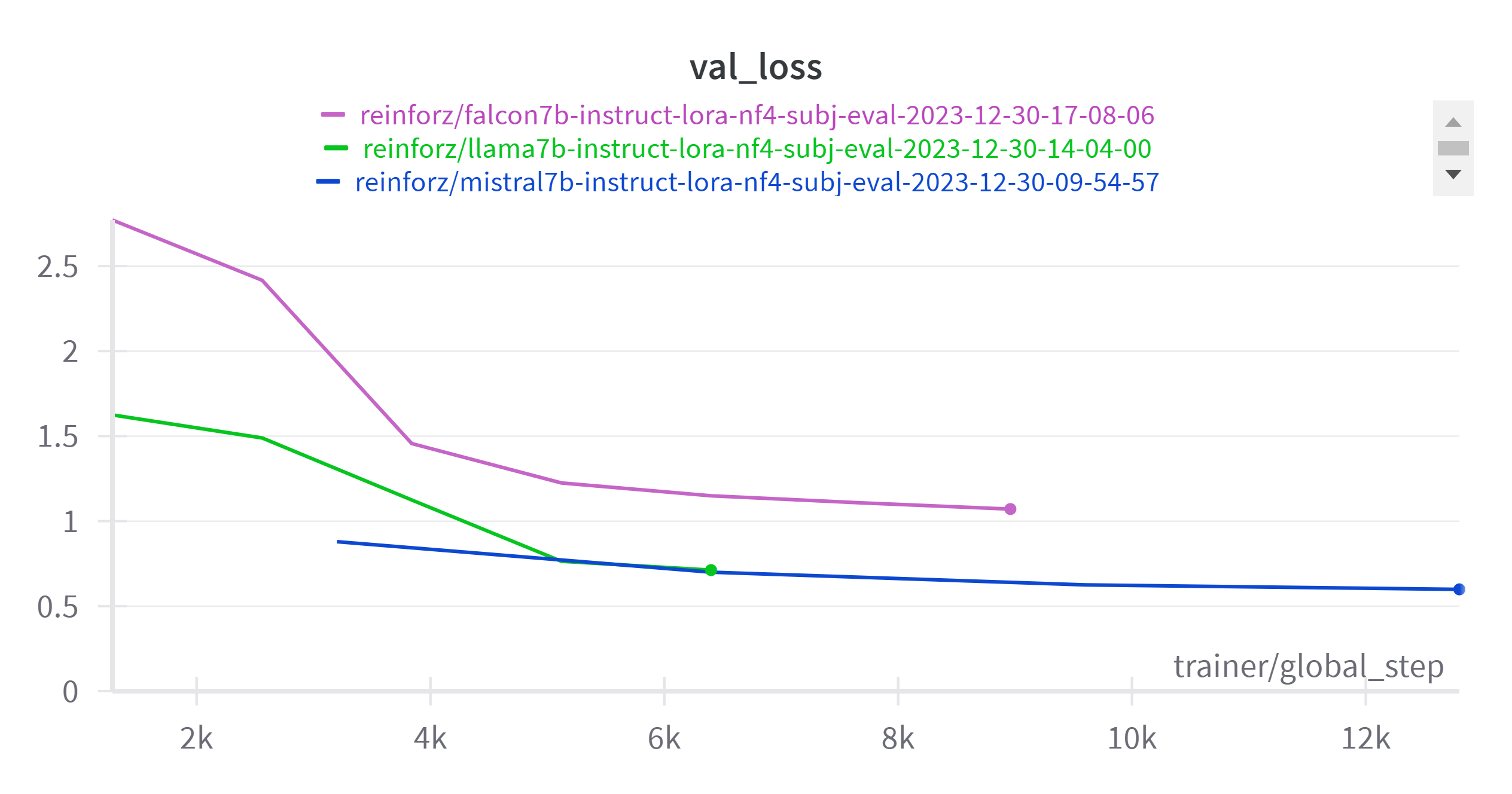}
    \caption{Train \& validation loss for answer evaluation model}
    \label{fig:eval-model-train-eval-loss}
\end{figure}

\begin{figure}
    \centering
    \includegraphics[width=.24\textwidth]{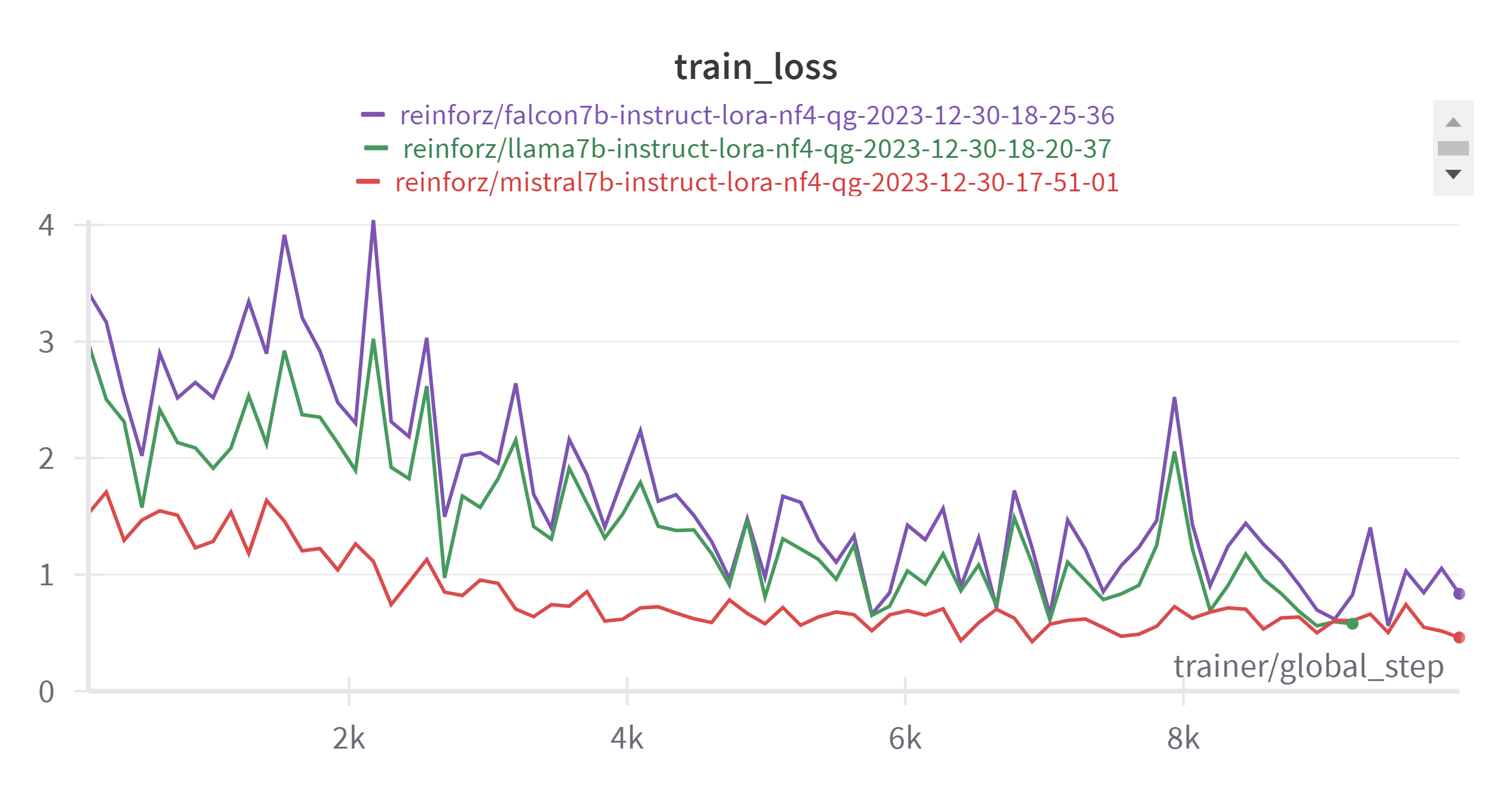}\hfill
    \includegraphics[width=.24\textwidth]{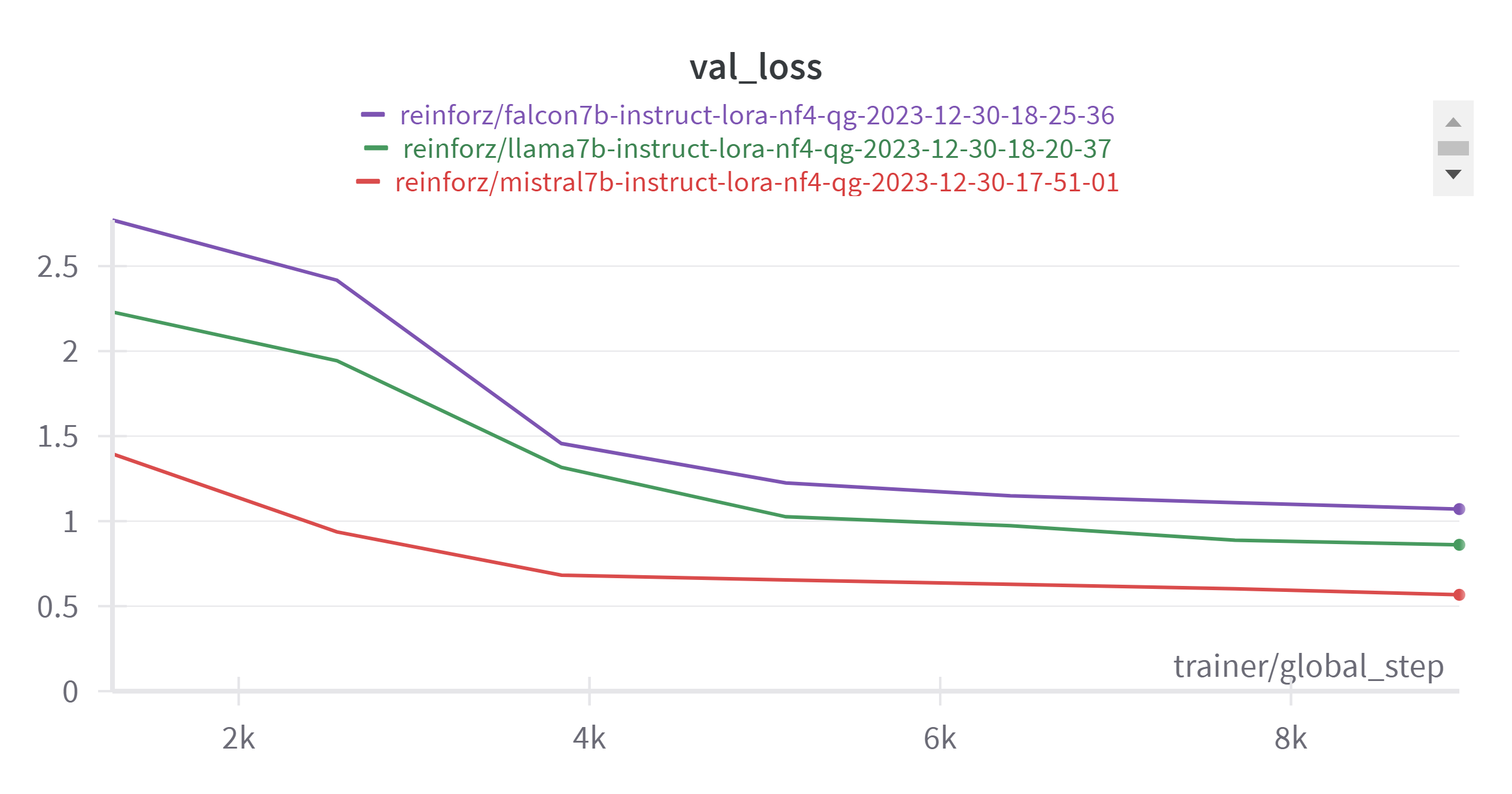}
    \caption{Train \& validation loss for question generation model}
    \label{fig:qgen-model-train-eval-loss}
\end{figure}

The training and validation loss for the subjective answer evaluation models and subjective question generation models were visualized as shown in Figure \ref{fig:eval-model-train-eval-loss} and Figure \ref{fig:qgen-model-train-eval-loss} respectively.

\section{Results}
\label{results}

\subsection{Model Evaluation Metric}
We used GPT-4 as the primary tool for model evaluation. Conventional Metrics, such as BLEU, ROUGE, etc., rely on matching a predetermined correct output. Since both the tasks, Subjective Question Generation and Answer Evaluation, are subjective by nature, they do not have a fixed correct answer and require a high-level understanding of natural language. Therefore, the aforementioned metrics are not suitable for our purposes. Human Evaluation can solve these issues but poses limitations in terms of cost and time. Furthermore, variations in the perspectives of different evaluators would lead to a lack of consistency in the evaluation of the models. According to Sottana et. al. \cite{sottana2023evaluation}, GPT-4’s evaluation aligns with Human Evaluation. So, with GPT-4, we can get a consistent human-like evaluation throughout the test dataset. As a result, we thought the model evaluation for the tasks of our research could be best done using GPT-4.

We ranked the model responses using GPT-4 for each test prompt from 1-4, with 1 being the best and 4 being the worst. Along with our three fine-tuned models, we also included the responses generated by GPT-3.5 as a benchmark for comparison. To avoid any biases, we anonymized the model names before feeding the model responses into GPT-4 for ranking. 

\subsection{Question Generation}
Table \ref{tab:question_generation_res} shows the aggregated results of GPT-4 rankings for Question Generation.

In Question Generation, Mistral 7B establishes itself as the best-performing model by generating top responses for over 65\% test prompts. GPT-3.5 is the second-best model and ranks top over 30\% of the time. Llama-2 7B and Falcon 7B show poor performance by largely ranking in the 3rd and 4th position, respectively. 

\begin{table}[ht]
\centering
\caption{Aggregated GPT-4 Rankings for Question Generation}
\label{tab:question_generation_res}
    \begin{tabular}{|l|r|r|r|r|}
        \hline
        Model/Rank &  1st &  2nd &  3rd &  4th \\
        \hline
        Mistral 7B &  65.78\% & 24\% & 7.11\% & 3.11\% \\
        \hline
        Llama-2 7B & 3.56\% &   7.56\% &  79.56\% & 9.33\% \\
        \hline
        Falcon 7B  & 0\% &  1.33\% &   11.11\% &  87.56\% \\
        \hline
        GPT-3.5 & 30.67\% & 67.11\% & 2.22\% &    0\% \\
        \hline
    \end{tabular}
\end{table}

\subsection{Answer Evaluation}
Table \ref{tab:answer_eval_res} shows the aggregated results of GPT-4 rankings for Answer Evaluation.

Here, GPT-3.5 shows the best performance with a 58.22\% top ranking. Mistral closely follows by ranking 1st in 40.44\% of the test cases. Llama-2 7B and Falcon show performance similar to their question generation performance. 

\begin{table}
    \centering
    \caption{Aggregated GPT-4 Rankings for Answer Evaluation}
    \begin{tabular}{|l|r|r|r|r|}
        \hline
        Model/Rank &  1st &  2nd &  3rd &  4th \\
        \hline
        Mistral 7B & 40.44\% &  48\% & 6.67\% & 4.88\% \\
        \hline
        Llama-2 7B & 1.33\% & 12\% & 76.89\% & 9.78\% \\
        \hline
         Falcon 7B &    0\% & 1.78\% & 12.89\% & 85.33\% \\
        \hline
        GPT3.5 &  58.22\% & 38.22\% & 3.55\% & 0\% \\
        \hline
    \end{tabular}    
    \label{tab:answer_eval_res}
\end{table}

\section{Discussion}
\label{discussion}

\begin{figure}
    \centering
    \includegraphics[width=.24\textwidth]{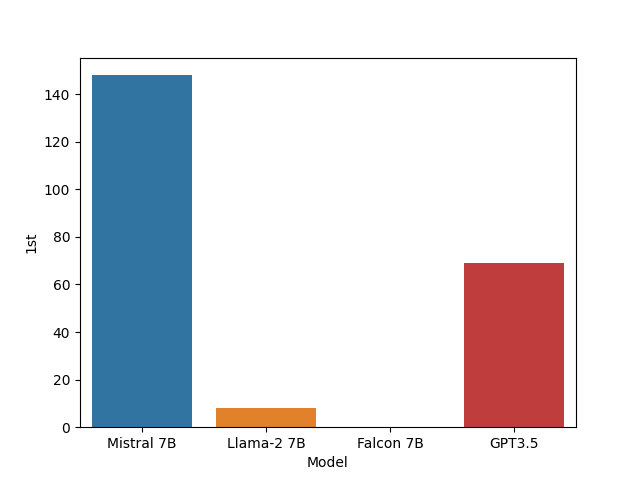}\hfill
    \includegraphics[width=.24\textwidth]{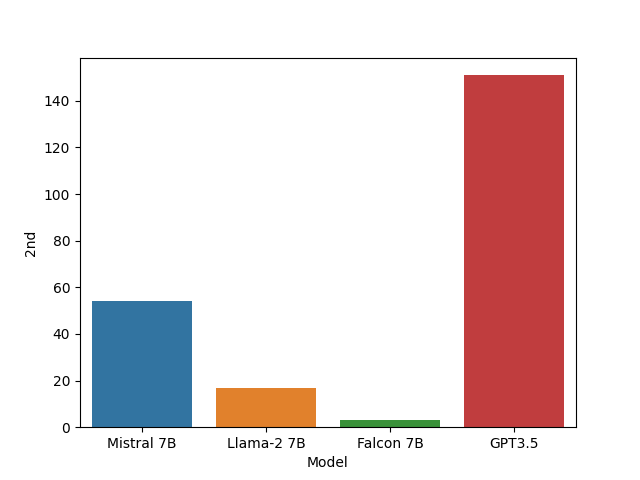}
    \\[\smallskipamount]
    \includegraphics[width=.24\textwidth]{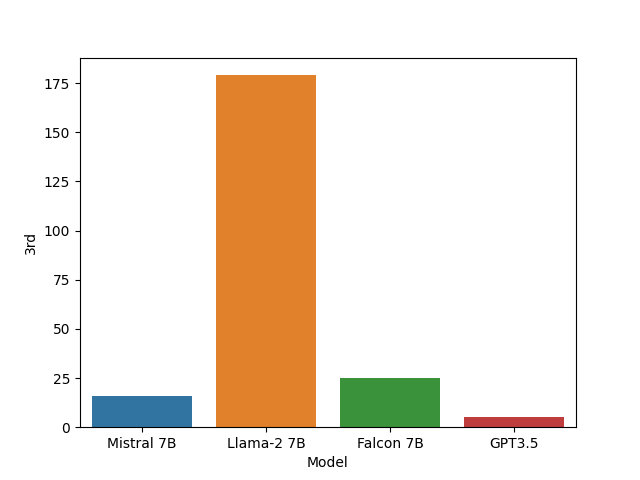}\hfill
    \includegraphics[width=.24\textwidth]{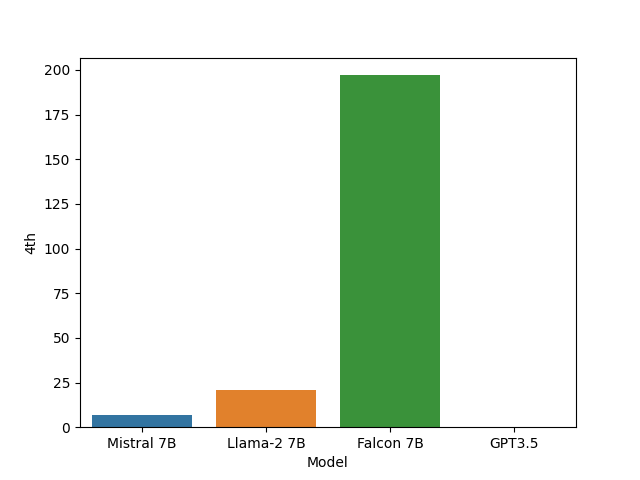}
    \caption{Distribution of Question Generation Model Placement for Each Rank}
    \label{fig:qg-barplot}
\end{figure}

Mistral 7B proves to be highly effective in terms of question generation by ranking top in the most number of test examples (65.78\%). It ranks in the top half almost 90\% of the time. This indicates that instruct-tuned Mistral 7B can produce high-quality subjective questions.

However, in terms of consistency, GPT-3.5 is ahead of Mistral 7B. Even though it ranks at the top less than Mistral, its generated responses are ranked 1st or 2nd almost 97\% of the time. Moreover, its responses were never ranked to be last for any test prompts. This highlights the consistent ability of GPT-3.5 to generate high-quality questions.

Llama-2 7B, after being fine-tuned, ranked predominantly at 3rd(79.56\%). This is consistent with the observation that Llama-2 7B adds extra irrelevant details after its generated questions. It only ranks in the top half 10\% of the time.

Falcon 7B demonstrated the weakest performance, predominantly ranking last (87.56\%). This aligns with the observation that it often generates gibberish or the worst responses. None of its responses ranked first, and its presence in the 2nd and 3rd positions is minimal at about 12\% 

In conclusion, after fine-tuning, Mistral 7B demonstrates a performance comparable to GPT-3.5 in terms of Question Generation quality. However, Llama-2 7B and Falcon 7B show poor performance. The poor performance of these models could be because of architectural differences in these models or due to the fact that they weren't optimized for these tasks. More research is needed to shed light on this.

\subsection{Answer Evaluation}

\begin{figure}
    \centering
    \includegraphics[width=.24\textwidth]{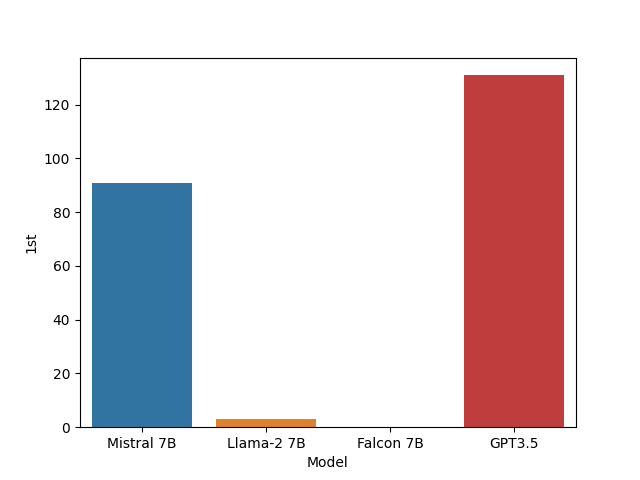}\hfill
    \includegraphics[width=.24\textwidth]{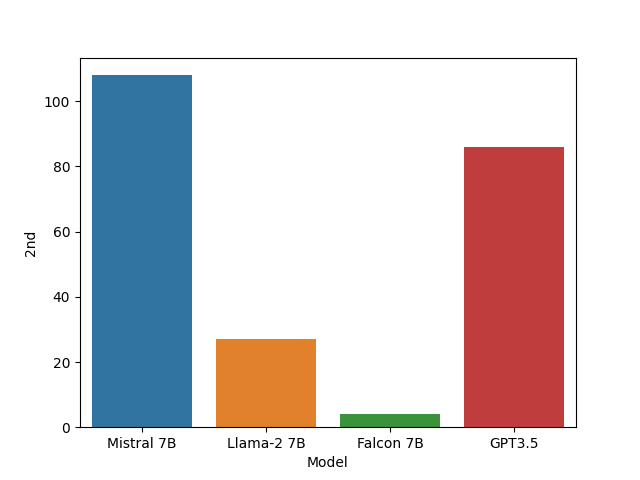}
    \\[\smallskipamount]
    \includegraphics[width=.24\textwidth]{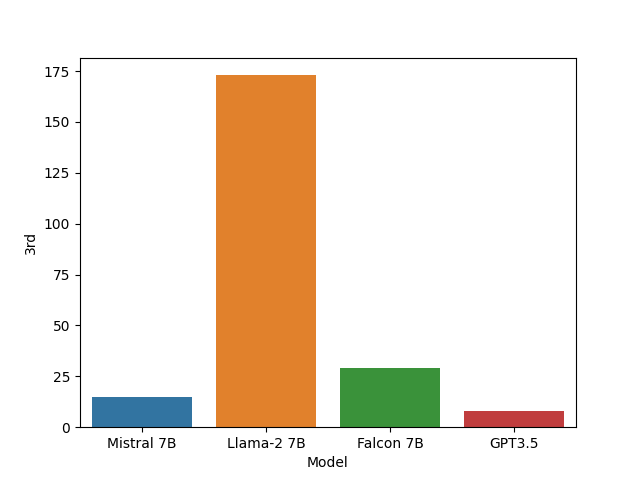}\hfill
    \includegraphics[width=.24\textwidth]{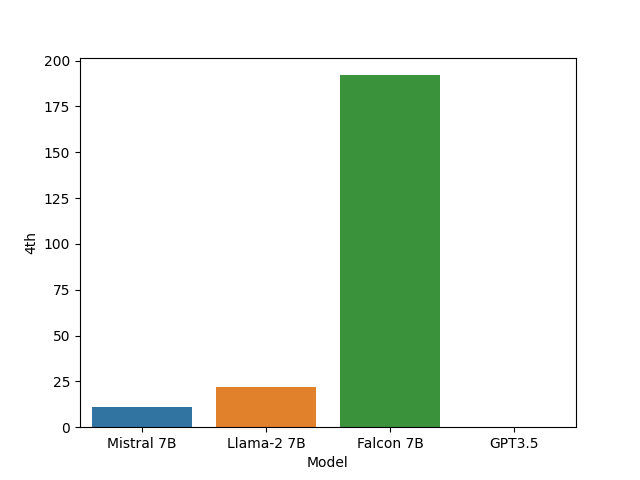}
    \caption{Distribution of Answer Evaluation Model Placement for Each Rank}
    \label{fig:eval-barplot}
\end{figure}

The results show that GPT-3.5 consistently outperforms other models. Its consistency in its generation of answer evaluation is made more evident because it ranks 1st or 2nd over 96\% of the time and does not rank last in test cases. Mistral 7B closely follows GPT-3.5 in the task of answer evaluation. It ranks 1st around 40\% 

Like Question Generation, Llama-2 7B and Falcon 7B fail in Answer Evaluation tasks, ranking predominantly in the bottom half. Llama-2 7B leads all other models by a mile in 3rd position. Its low percentage of ranking in 1st and 2nd place at around 13\% signifies its inability to generate a consistent and high-quality evaluation for the prompted answers. Falcon 7B again comes out at last for the most part and has no top rankings. Its ranking in 2nd and 3rd position is also negligible.

In summary, GPT-3.5 performs the best in Answer Evaluation, outperforming the finetuned models. Although Mistral 7B is not on par with GPT-3.5, its performance shows promise. Since Answer Evaluation is a much more complex task than Question Generation, GPT-3.5 performs better due to its large size than Mistral. In the future, larger models could be finetuned for the Subjective Question Answer Evaluation task to test if they can perform on par with GPT-3.5.

\section{Conclusion}
\label{conclusion}
In conclusion, this thesis has explored the use of Natural Language Processing in education and its potential to enhance how we learn, teach, and evaluate. We have highlighted the applications of LLMs in creating personalized learning experiences, generating and assessing questions, and evaluating students' writing.

The limitations of the research are also highlighted below, and the importance of further research in this field is emphasized. Future research can look into applying other NLP techniques or exploring other areas of study and improving current methodologies.

\section{Future Work}

\begin{itemize}
    \item Exploring other newer NLP techniques instead of just relying on LLMs for subjective question generation and answer evaluation
    \item Using more human-made data instead of synthetic ones.
    \item Exploring multiple other new open-source LLMs and performing comparative analysis on their generated output.
    \item Instruct-tuning larger or better versions of current or future models for this task.
    \item Using a multitude of evaluation metrics in order to assess the quality of the generated text. Involving humans in the loop to ensure the evaluation is as accurate as it can possibly be.
    \item Explore the changes in the accuracy of using various other finetuning or future instruct-tuning approaches.
\end{itemize}

\bibliographystyle{IEEEtran}
\bibliography{reference}
\vspace{12pt}
\color{red}
\end{document}